# Kannada Spell Checker with Sandhi Splitter

Akshatha A N
Department of ISE
RVCE, Bangalore

Chandana G Upadhyaya
Department of ISE
RVCE, Bangalore

Rajashekara Murthy S
Associate Professor, Department of ISE
RVCE, Bangalore

*Abstract*—Spelling errors are introduced in text either during typing, or when the user does not know the correct phoneme or grapheme. If a language contains complex words like sandhi where two or more morphemes join based on some rules, spell checking becomes very tedious. In such situations, having a spell checker with sandhi splitter which alerts the user by flagging the errors and providing suggestions is very useful. A novel algorithm of sandhi splitting is proposed in this paper. The sandhi splitter can split about 7000 most common sandhi words in Kannada language used as test samples. The sandhi splitter was integrated with a Kannada spell checker and a mechanism for generating suggestions was added. A comprehensive, platform independent, standalone spell checker with sandhi splitter application software was thus developed and tested extensively for its efficiency and correctness. A comparative analysis of this spell checker with sandhi splitter was made and results concluded that the Kannada spell checker with sandhi splitter has an improved performance. It is twice as fast, 200 times more space efficient, and it is 90% accurate in case of complex nouns and 50% accurate for complex verbs. Such a spell checker with sandhi splitter will be of foremost significance in machine translation systems, voice processing, etc. This is the first sandhi splitter in Kannada and the advantage of the novel algorithm is that, it can be extended to all Indian languages.

## I. INTRODUCTION

Kannada is an agglutinative language. It is one of the Dravidian languages, and by the nature of the Dravidian languages it has very clear rules defined for every aspect of its structure. Kannada has roughly 40 million native speakers and it is one of the 40 most spoken languages in the world [1]. It is influenced greatly by Sanskrit, and therefore we can find an overlap of words, structure and grammar rules including the sandhi and lexicon between the two languages. Like any other language, Kannada has grown and will continue to grow and change with the intervention of other languages and accents, and by people who want to make the language and its words easy to pronounce, spell and write. There is no specific boundary to the words in it. In a language like Kannada, where there are abundant complex structures and compound words, a spell checker demands a sandhi splitter for two reasons. First, since any database of Kannada words cannot store every sandhi word without huge redundancy, the sandhi splitter would hugely reduce the dictionary size. Second, sandhi splitters are critical for recognizing spelling errors arising due to an erroneous morpheme or an erroneous segment at the morpheme boundary of a sandhi word.

A morpheme is the smallest meaningful unit in a language. Joining morphemes to derive complex and meaningful words without changing the spelling or the phonetics of the constituent morphemes is called agglutination. Inflection, on the other hand, is the refitting of the words to express various grammatical aspects like gender, tense, mood and number.

In the processing of any language, morphological analysis, sentence structure analysis and recognition become the founding pillars. In processing Indian languages, in addition to the aforementioned factors, several factors such as sandhis, samaasas, and inflections specific to gender and tense also play a role. In Kannada, there are three ways of forming complex words: samaasa, jodi pada and sandhi.

*A. Samaasa, and Jodi Pada*

Samaasa is also known as nominal compound. Morphologically, a samaasa has each noun or adjective in its stem form with only the last element obtaining the case inflection. Examples of samaasa include "peetaMbara" and "vRukoodara". A jodi pada is a phonemic binding of two unrelated morphemes separated by a hyphen used in the Kannada dialect. Examples of jodi pada include "mane-maTha" and "deevaru-diMDaru".

*B. Sandhi*

Sandhi means 'to join'. In sandhi formation at the word boundary, several phonological processes take place to produce the complex word or the sandhi word. During this process of joining, one or both following operations occur at the word boundary:

- A new letter will appear at the word boundary.

- A letter will disappear from the end of the first word (prefix) or from the beginning of the second word (suffix) at the word boundary.

Sandhis have a set of definite rules that are defined in Panini's Ashtadhyayi [2]. Sandhis occur based on the last letter occurring in the prefix and the first letter occurring in the suffix. Based on the first letter occurring in the suffix, Panini classifies the sandhis as vowel (svara) and consonant (vyaMjana) sandhi in Sanskrit. Kannada language adapts all the sandhis in Sanskrit and has three additional sandhis that are confined just to Kannada. In this paper, the implementation of sandhi splitting of the four Sanskrit vowel sandhis namely savarNa deergha, guNa, vrddhi, yaN sandhi and three Kannada sandhis namely loopa, aagama and aadeesha sandhi, all for the Kannada language, is discussed, along with its application in a spell checker.

*C. Sandhi Splitter and Spell Checker*

A sandhi splitter is the one that separates the constituent words of a sandhi word using a comprehensive dictionary and sandhi rules. Since most words in a language get their meaning based on the context, sandhi splitting becomes context based at times and hence poses ambiguity in machine processing. In manual processing, based on the understanding of the language, this ambiguity can be resolved. Making a machine learn the context and to use the grammar rules accordingly is a major challenge.

As Kannada is a partial phonemic language, there is partial correspondence between graphemes and phonemes. This gives a huge scope for spelling mistakes. The usage of ottakshara and mahapRaNas also contribute to the errors in the spelling. This makes it necessary to have a spell checker for Kannada.

A spell checker is a piece of software that verifies if the word is spelt correctly and alerts the user in the case of spelling error. In an Indian language, a spelling error might occur in two forms of words, a sandhi word (compound word) or a non-sandhi word (root word).

In the case of a sandhi word, the spelling error can occur at three places.

- Prefix
- Suffix (inflection is also considered as the suffix in this paper)
- Sandhi place or the word boundary

As Kannada is mostly comprised of complex words, to make an efficient spell checker either these words must be split as per their sandhis to find the error or the dictionary must contain all the compound words to compare with the misspelt word. It is preferable to have a sandhi splitter within the spell checker and store only the morphemes for a non-redundant, effective and expandable dictionary.

## II. RELATED WORK AND BACKGROUND

French scientist Gerard Huet, has developed an online tool called the 'Sanskrit Reader Companion' to segment and tag simple Sanskrit phrases [3]. To segment the words, the tool employs simple de-concatenation which provides several outputs in most of the cases.

The University of Hyderabad, JNU-Hyderabad, IIIT-Hyderabad, Sanskrit Academy Hyderabad, Poornaprajna Vidyapeetha Bangalore, Rashtriya Sanskrit Vidyapeetha Tirupathi and JRR Sanskrit University Jaipur have together developed an online Sanskrit sandhi splitter. It can take input in six different encodings as a word. It not only performs sandhi splitting but also samaasa splitting. It also has a morphological generator and analyzer with transliteration [4].

A Sanskrit sandhi splitter has been developed by Sachin Kumar and Diwakar Mishra for vowel sandhis. The input is split at every point to output all possible combinations [5]. Likewise, Marathi Sandhi Splitter has also been developed by Joshi Shripad S that splits the sandhi using the sandhi rules [6]. Amba Kulkarni in her sandhi analyzer has developed a method where the input is split into two words using the sandhi rules and checked by the morphological analyzer. If the words aren't recognized, then the sandhi splitting is done recursively [7]. Uma Maheshwar Rao uses an external sandhi splitter in his Telugu spell checker tool which uses the Generate-Analyze- Evaluate approach [8].

## III. SANDHI RULES

*A. Sanskrit Sandhis*

- *Savarna Deergha Sandhi*: Paninian rule 'akaH savarnee deerghaHa' which means if the prefix ends with a/i/u/R and the suffix begins with aa/ii/uu/RR the sandhi word will have aa/ii/uu/RR respectively at the word boundary. E.g. deeva + aalaya = deevaalaya ( a+aa=aa)

- *Guna Sandhi*: Paninian rule 'aadguNaHa'. Guna sandhi is identified when prefix ends with a/aa and suffix begins with i/ii or u/uu and the sandhi word

contains ee/oo respectively at the word boundary. E.g. suurya + udaya = suuryoodaya (a+u=oo)

- *Vriddhi Sandhi*: Paninian rule 'Vriddhireechi'. Vriddhi sandhi is identified when the prefix ends with a/aa and suffix begins with e/ee or o/oo and the sandhi word contains ee/ai respectively at the word boundary. Eg. eeka + eeka =eekaika (a+ee=ai)

- *Yan Sandhi*: Paninian rule 'ikooyaNaci' - Yan sandhi is identified when prefix ends with i/ii or u/uu and suffix begins with any vowel other than i/ii resulting in y/v induced at the word boundary of the sandhi word. E.g. manu + aMtara = manvaMtara ( u+a=v)

*B. Kannada Sandhis*

- *Loopa Sandhi*: Loopa sandhi is identified when the last vowel is omitted from the prefix at the word boundary with no restrictions on the beginning or the ending letters of the two native Kannada prefix and suffix words. E.g. uuru + uuru = uuruuru

- *Aagama Sandhi*: Aagama sandhi is identified when the letter 'y' or 'v' is introduced at the word boundary of two native Kannada prefix and suffix words. E.g. mara + annu = maravannu

- *Aadeesha Sandhi*: Aadeesha sandhi is identified when the letters 'k', 't', 'b' in the first position of the suffix is replaced by 'g', 'd', 'b' respectively in a sandhi word. E.g. maLe + kaala = maLegaala

## IV. DICTIONARY AND MORPHOLOGICAL ANALYZER

A dictionary that contains only the root words and non-compound words is necessary and sufficient for the spell checker with sandhi splitter described in this paper. A list which contains all possible inflections that verbs and nouns can take in the decreasing order of their priority is required to identify and remove the inflection so that the word in its native form is obtained.

*A. Directed Acyclic Word Graph (DAWG)*

A Directed Acyclic Word Graph, or DAWG, provides dictionary-like read-only objects. String data in a DAWG takes 200 times less memory than in a standard dictionary while maintaining a comparable raw lookup speed [9]. It also provides fast advanced methods like prefix search. These features of DAWGs are highly appreciated and useful for the implementation of the spell checker with sandhi splitter.

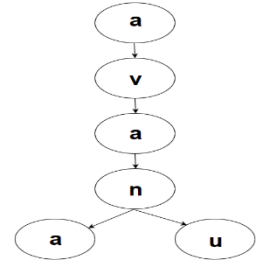

Fig.1. DAWG for avana and avanu

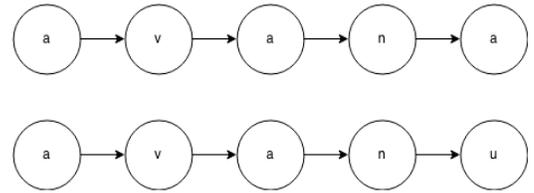

Fig. 2. Linear dictionary storage for avana and avan*u*

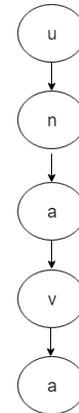

Fig. 3. Reverse DAWG of avanu

In a method where prefix word is read from left to right and the suffix is read from right to left, a reverse DAWG structure would be helpful. A reverse DAWG is obtained by reverting the word and constructing a DAWG structure of this reversed word. Since DAWG supports fast prefix search, a reverse DAWG provides fast methods to search suffix.

*B. Morphological Analyzer*

A vibhakti pratyaya is a case marker which consists of a letter or a bunch of letters attached to the end of a word. The morphological analyzer discussed in this paper can identify all the seven vibhakti pratyayas in Kannada along with the gender and plural markers. A dissected Kannada word will always be in following form:

[root word][gender marker][plural marker][pratyaya]

In morphological analysis, the word is scanned from the right till a valid vibhakti pratyaya is found. Though the main focus is on the nouns, a similar approach can be used to build an exhaustive morphological analyzer for all forms of verbs. There is a chance of occurrence of the sandhis at the joining of the root word and pratyayas. The sandhi formed is split and the valid pratyaya is removed successively. Similarly, multiple pratyayas, gender and plural markers are removed until only root word is left.

E.g. deevaalayagaLalli

alli: vibhakti pratyaya

gaLu: plural marker

After removal: deevalaya = root word.

## V. DESIGN AND IMPLEMENTATION

The concept of "ottakshara", which is one consonant immediately following another without an intervening vowel in Kannada makes it hard to process the Kannada text in Unicode format. Hence Romanization is necessary. There are several ways of Romanization and in this paper the Romanization technique proposed by Prof. Kavi Narayana Murthy has been adopted [10]. A one to one mapping of Kannada letters to a homophonic roman letter or a pair of letters is defined.

Fig.4. Romanization Chart

### A. Sandhi Splitter

Sandhi splitting can be done in the following two ways.

*1) Sandhi Place Determination and evaluation:* A sandhi is most likely to be occurring in positions of a word where deergha or ottakshara is found. Two consecutive vowels constitute a deergha and two consecutive consonants constitute an ottakshara. Such positions in a word are identified. By using sandhi rules, the possible splits are generated and evaluated for validity. Using a reverse sandhi rule base in the sandhi word, the sandhi letters are replaced with its corresponding constituent letters and split at that position to generate a prefix and a suffix. Prefix and suffix thus generated are searched in the dictionary for validity. If not found in the dictionary, there could either be a spell mistake in the words or it could be an invalid split. If the prefix and the suffix are found in the dictionary, then the split is valid. The correct sandhi would be the one whose rule was applied to get the split. An iterative search is made until a valid suffix or prefix is found.

E.g. suuryaoodaya

The two deerghas in the word are identified:
s(uu)rya(oo)daya
According to the savarnadeergha sandhi rule,
uu = u + u. By substituting and splitting,
suuryoodaya = su + uryoodaya. Prefix su and suffix uryoodaya are both not found in the dictionary.

According to the guNa sandhi rule, oo = a + u. By substituting and splitting, suuryoodaya = surya + udaya. Prefix surya and suffix udaya are both found in the dictionary. Therefore, the sandhi here is guNa.

*2) The Prefix-Suffix Method:* The given compound word is scanned from the left to find the prefix, which is looked up in the dictionary DAWG and the word is scanned from the right to find the suffix, which is looked up in the reverse DAWG. Using this approach, the longest valid prefix and suffix for the word are found. The resultant sandhi letters which are defined in the sandhi rules, if found at the overlapping letters of the prefix, sandhi rules are applied to generate a split.

E.g. deevaalaya

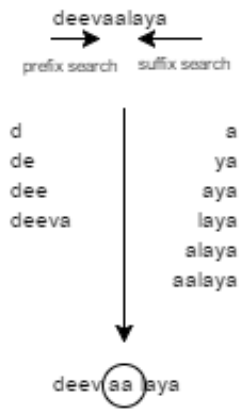

Fig 5. Prefix- Suffix Method

A novel approach which uses both sandhi place determination method and the prefix-suffix method could be used to enhance the results. The sandhi word given as an input is scanned from the left to right to find the longest prefix. This longest prefix will be referred to as expected prefix. This expected prefix is removed from the sandhi word leaving behind the sandhi letters and expected suffix, referred to as remainder word. The last letter of the expected prefix is then removed from the expected prefix and added to the beginning of the remainder word. The first one or two letters of the remainder word is most likely to be containing the resultant sandhi letters. These letters are looked up in the sandhi rules to identify the sandhi. Using the reverse sandhi rule base, the sandhi letters are replaced with the prefix's ending letter and suffix's beginning letter according to the sandhi rules. The expected prefix is then added to the remainder word and the words are split. The prefix and suffix thus generated are looked up in the dictionary. If both prefix and suffix are found, the sandhi rule which was applied to split the words is the required sandhi and the process is terminated as the sandhi, prefix and suffix words are determined successfully. If the sandhi is not determined, the second longest prefix is assigned as the expected prefix and the process is continued until the sandhi is determined or the expected prefix is null.

```
// Sandhi splitter pseudo code
sandhi_splitter(expected_prefix, remainder_word)
    while expected_prefix is not null and sandhi_found is not true
        set sandhi_letters to the first two chars of remainder_word
        set prefix_suffix_letters = search_reverse_sandhi_rules(sandhi_letters)
        // prefix_suffix_letters are of the form a+aa
        set sandhi_word = expected_prefix + sandhi_letters + remainder_word
        set sandhi_found = validate_sandhi(sandhi_word)
        set expected_prefix to the second longest prefix
        set remainder_word = remove_expected_prefix( given_sandhi_word )
        add the last letter of expected_prefix to remainder_word
        set expected_prefix = remove_last_letter(expected_prefix)
    endwhile
    if sandhi_found is true then
        return prefix, suffix and sandhi
    else
        return sandhi could not be found
    endif
```

Fig 7. Pseudo code of the novel approach

### B. Spell Checker

The spell checker with sandhi splitter checks the given word for spell errors and generates suggestions for misspelt words. The spelling errors in a corpus are mostly accidental and, usually, just one or two letters in a word are affected [11]. The Levenshtein distance is used to generate a list of suggestions for a misspelt word. The Levenshtein distance between two words is the minimum number of single-character edits like insertions, deletions or substitutions required to change one word into the other [12]. The spell checker with sandhi splitter can spell check and generate suggestions for all the words with single character errors.

The two main scenarios for spell checker with sandhi splitter are:

1) *Word is found in the dictionary*: If the word is found in the dictionary, the word is spelled correctly.
2) *Word is not found in the dictionary*: If the word is not found in the dictionary, then either the word is a sandhi word or the word is misspelt.
    a) *The word is a sandhi word*: The sandhi splitter would split the word to find a valid sandhi based on sandhi rules. If a sandhi split results in valid prefix and suffix, then the word is valid.
    b) *The word is misspelt*: If the word is not found in the dictionary and the sandhi split does not result in a valid prefix and suffix, the word is misspelt.

### C. Generating Suggestions

In case of a misspelt root word, all the words in the dictionary which are at a Levenshtein distance one are listed as suggestions. If the misspelt word is a sandhi word, then the suggestions are generated based on the position of the error and the error might be in one of the three positions:

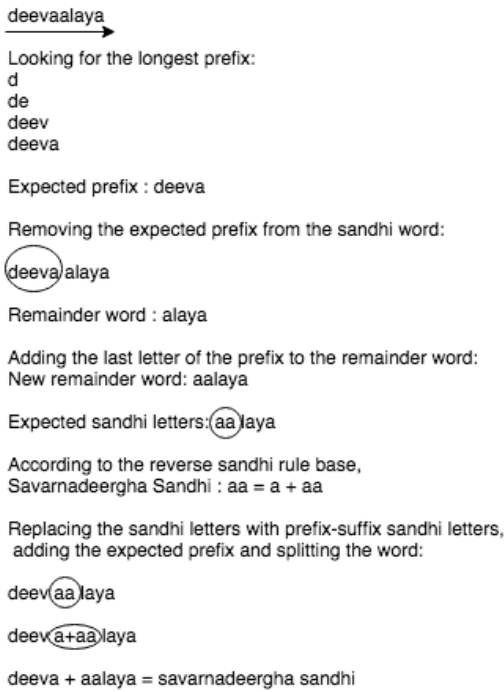

Fig. 6. Novel approach example

1) *Spell error in the suffix*: The sandhi splitter identifies the sandhi letters using an expected prefix. The sandhi is identified and its rules are applied at the sandhi point based on sandhi letters. A valid split would generate a valid prefix and a misspelt suffix. All the words starting with the suffix beginning letter given by the sandhi rules and at a Levenshtein distance of one to the expected suffix would be the intended suffixes. Each of them are joined with the prefix using the same sandhi rules with which it was split and a list of suggestions is generated.
2) *Spell error in the prefix*: Locating the sandhi letters becomes tedious if the prefix is misspelt. In such cases, the sandhi splitter looks for the expected suffix and sandhi letters are located. Based on the sandhi letters, the sandhi is identified and its rules are applied at the sandhi position. The sandhi split results in a misspelt prefix and a valid suffix. All the words ending with the same letter as the prefix ending letter given by the sandhi rules and are at a Levenshtein distance of one to the expected prefix would be the intended prefixes. Each of them are joined with the suffix using the same sandhi with which it was split and a list of suggestions is generated.
3) *Spell error the sandhi letters*: Since locating the sandhi letters is impossible in this case, expected prefix and suffixes are identified. Using the ending and beginning letters of the expected prefix and suffix, all possible expected sandhis are identified. Sandhi rules of each expected sandhi is applied to get the intended sandhi words. All the intended sandhi words which are at a Levenshtein distance of one to the misspelt word are added to the list of suggestions.

The spell checker with sandhi splitter also stores the last picked suggestion and shows it on top the next time same error appears. Along with spell checking and generating suggestions, the spell checker with sandhi splitter can also be helpful in:

*a) Root word extractor*: The words containing pratyayas are most likely to have a sandhi formed between the pratyayas and the root word. Using the sandhi splitter, the root word is extracted and the pratyayas are separated.
*b) Semantic analyser*: The semantic analysis of the word could be made based on the gender, number and tense. These indicators on the features mostly contain a sandhi word. Using a sandhi splitter and spell checker, semantic analysis of the words could be made.
*c) Samasa analyser*: Though it is not necessary for a samaasa word to have a sandhi, in the cases where a sandhi is found in a samaasa word, the sandhi splitter can be used to find root words.

## VI. EXPERIMENT AND RESULTS

The time and space efficiency and performance of the spell checker with sandhi splitter is measured and is compared to other conventional spell checkers. The DAWG structure used in the spell checker with sandhi splitter for dictionary is highly space efficient and it takes 200 times less space than a spell checker with linear storage structure.

The DAWG also provides advanced methods for prefix search. The spell checker with sandhi splitter has been tested with a test sample of around 700 nouns and verbs and its various inflections. A verb takes about 80 inflections whereas a noun takes around 10 inflections. The verbs change drastically when it takes an inflection making it difficult for the spell checker with sandhi splitter. The spell checker with sandhi splitter has an improved performance of 90% correct result in the case of nouns and about 50% in the case of complex verbs compared to a conventional spell checker which has only 60% efficiency with only root words stored in the dictionary. Clearly, the sandhi splitter used in the spell checker validates the sandhis and provides with an enhanced performance. The performance can be enhanced further by extending the sandhi splitter to split consonant sandhis. The graphs from the experiment clearly show that the spell checker with sandhi splitter performs better than the conventional spell checker.

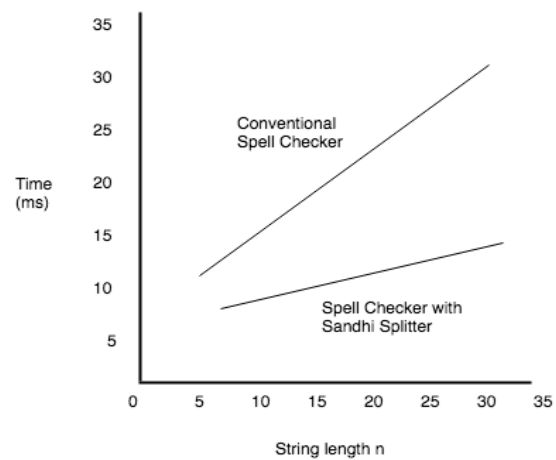

Fig. 8. Time efficiency of spell checker with sandhi splitter compared to that of a conventional spell checker

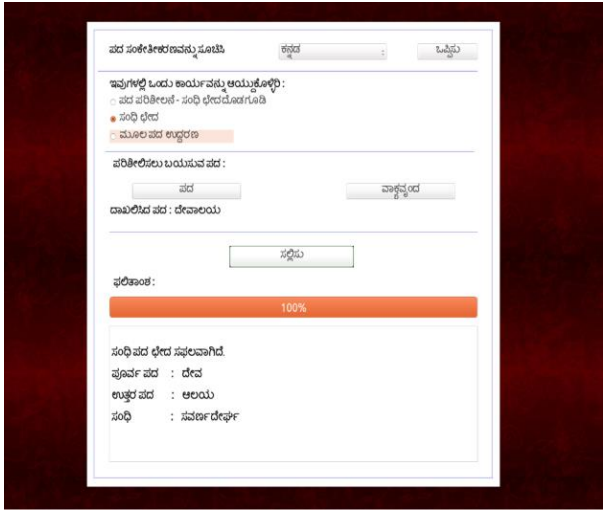

Fig. 9. Sandhi Splitter splitting a Kannada Sandhi word

This Spell Checker with Sandhi Splitter was developed with Kannada and English language script support. Spell Checker with Sandhi Splitter, Sandhi Splitter and Root Word Extractor are the three functions of this application. This tool is capable of taking either a single word or an entire corpus as an input. A simple semantic analyzer developed using this application also showed convincing results. Providing to the user an option to add new words to the dictionary greatly increased the performance of the application.

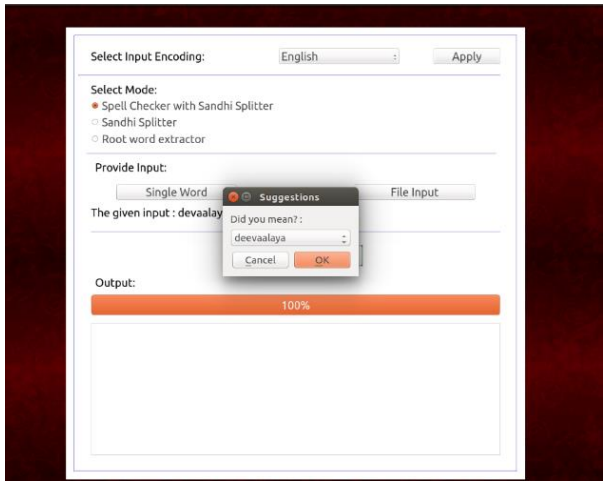

Fig. 10. Suggestions generated by spell checker with sandhi splitter

## VII. CONCLUSION

The spell checker with sandhi splitter achieves overall 80% accuracy with an efficient dictionary. Since Kannada has a lot of complex words, a sandhi splitter boosts the performance of the spell checker by a considerable amount. The novel algorithm proposed could be extended to split consonant sandhis. Using the dictionary and sandhi rules of any Indian language, the application will be able to spilt sandhis of that language. The Spell Checker with Sandhi Splitter can be improved upon further to provide a list of exhaustive suggestions by taking into consideration the semantics and the context of the word. The sandhi splitter can also be used as a root word extractor in machine translations systems and can be used as a basic tool for semantic analysis. The tool can be enhanced to identify complex samasa words and resolve them. This application will be helpful in solving many Natural Language Processing problems efficiently.